# Can ChatGPT Learn to Count Letters?

Javier Conde, ETSI de Telecomunicación, Universidad Politécnica de Madrid, 28040 Madrid, Spain

Gonzalo Martínez, Universidad Carlos III de Madrid, 28911 Leganés, Madrid, Spain

Pedro Reviriego, ETSI de Telecomunicación, Universidad Politécnica de Madrid, 28040 Madrid, Spain

Zhen Gao, Tianjin University, 300072 Tianjin, China

Shanshan Liu, University of Electronic Science and Technology of China, 611731 Chengdu, China

Fabrizio Lombardi, Northeastern University, Boston, MA 02115, USA

**Abstract**—Despite their amazing capabilities on many different tasks, Large Language Models (LLMs) struggle on simple tasks such as counting the number of occurrences of a letter in a word. In this paper we explore if ChatGPT can learn to count letters.

Since the introduction of ChatGPT two years ago, Large Language Model (LLM) based tools have shown impressive capabilities to solve mathematical problems or to answer questions on almost any topic [1]. In fact, evaluation benchmarks have to be revised frequently to make then harder as LLM performance improves continuously [2].

The development of LLMs has also been hectic with new models presented by large companies such as Google with Gemini or Gemma, Meta with Llama or x.AI with Grok. OpenAI has also released newer versions and improvements of their Generative Pre-trained Transformer (GPT) family such as GPT4 [3] and its variants GPT4o and GPT4o1. Those foundational models are then adapted to answer questions or interact with users and complemented with other functionalities to implement conversational tools like ChatGPT.

Despite these astonishing results, there are some simple tasks that LLMs struggle with, for example arithmetic operations [4] or even counting the occurrences of a given letter in a word. For example, many LLMs failed to count the number of "r" in strawberry[1]. The failure to perform this task stems from how LLMs are designed as they internally they do not work with letters.

Off the shelf LLMs can be fine-tuned using examples to adapt then to a given application or to produce the output in a given format [5]. Therefore, an interesting question is whether LLMs can learn to count letters by fine-tuning the models. In this paper we explore this idea and show how developing it provides insights into how LLMs work and learn. The starting point is to understand how LLMs operate and why they can fail to count letters.

## WHAT ARE LETTERS? LLMs ONLY KNOW ABOUT TOKENS

For humans, the basic element of a language are its letters that we learn in primary school and that we use subsequently to form words and to learn the language. This means that we are able to identify and count letters from an early age. Instead, LLMs do not have the concept of a letter, they operate on units called tokens.

A token is a sequence of letters that appears frequently in the text, for example let us consider the word "strawberry" and the tokenizer of GPT4o[2]. This tokenizer splits the word in three "str", "raw" and "berry" tokens as shown in Figure 1 that includes two additional examples of the tokenization of words. It can be seen that tokens do not necessarily correspond to words, syllables or letters.

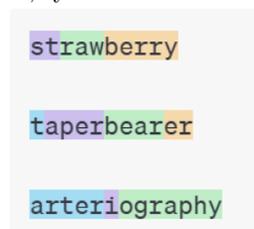

FIGURE 1. Examples of word tokenization in GPT4o

To understand how tokens are constructed we have first to understand how LLMs work. LLMs simply predict the next token based on the previous text. Then use the previous text and the generated token as the input to the model and so on. This is illustrated in Figure 2 for a simple prompt. The LLM generates the first token "The" and then uses the initial prompt plus this token to ask the LLM to predict the next

---

[1] https://medium.com/@davehk/llms-the-strawberry-riddle-technical-background-65bed7444ede
[2] https://platform.openai.com/tokenizer

token "capital" and so on. In this example, each token corresponds to a word.

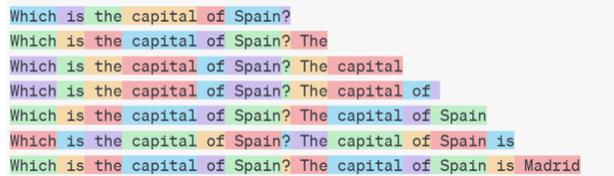

FIGURE 2. Example of a sequence of next token predictions

The process of generating text is therefore done sequentially token by token. This means that if we were to use letters, generating the answer to that prompt would require 25 predictions, instead in the example using GPT4o we only need to predict 6 tokens. This is a very significant reduction specially as the prediction of each token has a large computational cost in LLMs. However, using letters instead of tokens has also advantages. An obvious one, is that the LLM has fewer options to choose from for each prediction which reduces the complexity of the prediction as LLMs have to store and process information for each possible prediction outcome. This leads to a design trade-off between the number of distinct tokens and the number of predictions needed to answer to a prompt. The more distinct tokens the fewer the predictions but each being more complex.

As in most engineering problems, the optimum value for the number of distinct tokens is not at the extremes: using letters requires too many predictions and having a token for each word requires too many distinct tokens. Current LLMs have tens or hundreds of thousands of distinct tokens, for example GPT4o has approximately 200K distinct tokens. Since the goal is to reduce the number of predictions needed to answer a prompt, tokens are assigned to letter sequences that appear frequently in the text. This means that, if a word is very common even if its long, it is likely that a token will be assigned to it unless it is made of common sequences of letters. For example, "girlfriend" is mapped to two tokens: "girl" and "friend" but "measurement" is mapped to a single token "measurement" and "organizing" to "organ" and "izing". Tokens do not have to map to words, prefixes or any other language unit, the key to assign tokens is the frequency of the sequence and of other related sequences that contain it. The tokenizer that decomposes the input text into tokens is a key element of the LLM and its design involves complex algorithms that are commonly designed to map tokens to words or subword units [6].

The use of tokens means that LLMs do not operate with letters which makes it hard for them to count something they do not use. For example, to count the number of "r" in "strawberry" the LLM would have to know that tokens "str", "raw" and "berry" have 1,1 and 2 "r" respectively. Clearly by design LLMs will find hard to count letters. To check that is the case we have randomly selected 7,790 words from a public list of English words[3] with an even distribution of the number of letters per word and asked GPT4o to count the number of "r" letters on each of the words. The results are summarized in Table 1. It can be observed that even a complex and powerful LLM like GPT4o has a significant failure rate when counting letters.

TABLE 1. Results of counting the number of "r" in 7,790 words

| Correct | 6602 |
|---|---|
| Incorrect | 1188 |
| Failure rate (%) | 15.25% |

## CAN WE TEACH LLMs TO COUNT LETTERS

One possibility to have LLMs count letters is to fine-tune the existing models to this task. As discussed before fine-tuning adjusts the model parameters using a much smaller dataset that the one using used to train the model [5]. In our case, the dataset could be a few thousand words for which we have counted the number of "r" letters, so we provide the word and the number of "r" in it to fine-tune the model.

In order to evaluate the effectiveness of fine-tuning, another subset of 7,790 words is selected, so having two such subsets of words. Then the first subset is used to fine-tune the model to learn to count the number of "r" in each word and the second subset to measure the failure rates for both the original and the fine-tuned models. The results are summarized in Table 2. It can be observed that the fine-tuned model has a much lower failure rate, close to zero. Instead the original GPT4o has a much larger failure (same as in Table 1). Therefore, LLMs can learn to count letters by fine-tuning. Since LLMs work with tokens, during the fine-tuning process, it seems that the model learns how many "r" are in each token so that given a sequence of tokens that correspond to a word it can infer the number of "r" in that sequence.

TABLE 2. Results of counting the number of "r" in the 7,790 test words for the original and fine-tuned models

| Model | GPT4o | Fine-tuned |
|---|---|---|
| Correct | 6602 | 7733 |
| Incorrect | 1188 | 57 |
| Failure rate (%) | 15.25% | 0.74% |

Let us consider a person that is taught how to count "r" letters in a word. If later we ask that person to count the number of "a" letters or the number of "m" letters in a word, the person will be able to count those letters too, being able to generalize as she has learned how to count letters in general, not only "r". Would the same hold for LLMs? Will the fine-tuned model reduce the failure rate when counting other letters?

To try to answer that question, the original GPT4o and the version fine-tuned to count "r" letters have been asked to count the number of "a" and "m" letters in the 7,790 test words.

---
[3] https://github.com/dwyl/english-words/blob/master/words_dictionary.json

observed that the failure rates are reduced also when counting "a" and "m" letters on the model fine-tuned to count "r" letters. This suggest that to some extent the LLM is learning how to count letters in general and not only "r", similarly to what persons do.

TABLE 3. Results of counting the number of "a" and "m" in the 7,790 test words for the original and fine-tuned to count "r" letters models

|  | Counting "a" | | Counting "m" | |
|---|---|---|---|---|
| Model | GPT4o | Fine-tuned | GPT4o | Fine-tuned |
| Correct | 6779 | 6807 | 6873 | 7665 |
| Incorrect | 1011 | 983 | 917 | 125 |
| Failure rate (%) | 12.98 | 12.62 | 11.77 | 1.60 |

To better understand how well the fine-tuned model generalizes, we have run a third experiment in which we fine-tune GPT4o to count the number of "a" and "m" letters respectively and compare the results of the fine-tuned models when used to count the letter they were fine-tuned for and the other two letters. The results are summarized in Table 4 and show that the fine-tuned models are able to generalize for other letters and not only for the one fine-tuned, except for the vowel "a" where results are similar for the base model and the other letters finetuned models. Therefore, it seems that in the fine-tuning the model is not only learning to count the letter used for training but also to some extent is learning to count other letters. This capability of LLMs to generalize shows the potential of LLMs to adapt to new tasks by fune-tuning.

TABLE 4. Failure rates (%) of the models in the 7,790 test words when counting the three letters

| Model | Counting "r" | Counting "a" | Counting "m" |
|---|---|---|---|
| GPT4o | 15.25 | 12.98 | 11.77 |
| Fine-tuned for "r" | 0.74 | 12.61 | 1.60 |
| Fine-tuned for "a" | 8.42 | 0.44 | 4.85 |
| Fine-tuned for "m" | 5.28 | 13.00 | 0.55 |

## LEARNING FROM COUNTING LETTERS

The analysis of a simple problem, why LLMs fail to count the "r" letters in a word, has served us to illustrate some important aspects of how LLMs process text and also how they can learn to perform specific tasks with fine-tuning. Interestingly, this simple problem also illustrates the capabilities of LLMs to generalize beyond what they have been fine-tuned for, similar to what happens with persons.

The problem of counting letters has no interest from a practical point of view, as a traditional computer program can count letters much more efficiently and some LLM based tools have also been adapted to be able to count letters. In fact, some newer models like GPT4o1 (apparently named internally "strawberry" prior to its release) provide advanced reasoning capabilities that can solve this and much more complex problems. However, analyzing why LLMs can fail to count letters and how they can learn to do it provides an intuitive and simple case study to help understand some of the limitations and capabilities of LLMs.

## ACKNOWLEDGMENTS

This work was supported by the Agencia Estatal de Investigación (AEI) (doi:10.13039/501100011033) under Grant FUN4DATE (PID2022-136684OB-C22), by the European Commission through the Chips Act Joint Undertaking project SMARTY and by the OpenAI API Researcher Access Program.

Javier Conde is an assistant professor at the ETSI de Telecomunicación, Universidad Politécnica de Madrid, 28040 Madrid, Spain. Contact him at javier.conde.diaz@upm.es

Gonzalo Martínez is a researcher at Universidad Carlos III de Madrid, 28911 Leganés, Madrid, Spain. Contact him at gonzmart@pa.uc3m.es

Pedro Reviriego is an associate professor at the ETSI de Telecomunicación, Universidad Politécnica de Madrid, 28040 Madrid, Spain. Contact him at pedro.reviriego@upm.es

Zhen Gao is an associate professor at Tianjin University, 300072 Tianjin, China. Contact him at zgao@tju.edu.cn

Shanshan Liu is a professor at the University of Electronic Science and Technology of China, Chengdu 611731, Sichuan, China. Contact her at ssliu@uestc.edu.cn

Fabrizio Lombardi is a professor at Northeastern University, Boston, MA 02115, USA. Contact him at lombardi@coe.northeastern.edu